\documentclass[11pt]{article}
\usepackage[ruled,vlined]{algorithm2e}
\SetKwInput{KwIn}{Input}
\SetKwInput{KwOut}{Output}
\usepackage{amssymb}
\usepackage{placeins}

\usepackage[preprint]{acl}

\usepackage{times}
\usepackage{latexsym}

\usepackage[T1]{fontenc}

\usepackage[utf8]{inputenc}

\usepackage{microtype}

\usepackage{inconsolata}

\usepackage{graphicx}
\usepackage{amsmath}
\usepackage{color}
\usepackage{colortbl}
\usepackage{booktabs}
\title{FastV-RAG: Towards Fast and Fine-Grained Video QA with Retrieval-Augmented Generation}

\author{Gen Li \\
  University of Electronic Science\\
  and Technology of China\\
  Chengdu, China \\
  \texttt{leog3n@gmail.com} \\\And
  Peiyu Liu\thanks{Corresponding author.} \\
  University of International Business\\
  and Economics \\
  Beijing, China \\
  \texttt{liupeiyustu@163.com} \\}

\begin{document}
\maketitle
\begin{abstract}
Vision–Language Models~(VLMs) excel at visual reasoning but still struggle with integrating external knowledge.
Retrieval-Augmented Generation~(RAG) is a promising solution, but current methods remain inefficient and often fail to maintain high answer quality.
To address these challenges, we propose \emph{VideoSpeculateRAG}, an efficient VLM-based RAG framework built on two key ideas.
First, we introduce a speculative decoding pipeline: a lightweight draft model quickly generates multiple answer candidates, which are then verified and refined by a more accurate heavyweight model, substantially reducing inference latency without sacrificing correctness.
Second, we identify a major source of error—incorrect entity recognition in retrieved knowledge—and mitigate it with a simple yet effective similarity-based filtering strategy that improves entity alignment and boosts overall answer accuracy.
Experiments demonstrate that VideoSpeculateRAG achieves comparable or higher accuracy than standard RAG approaches while accelerating inference by approximately $2\times$.
Our framework highlights the potential of combining speculative decoding with retrieval-augmented reasoning to enhance efficiency and reliability in complex, knowledge-intensive multimodal tasks.
The codes are available at \url{https://github.com/FastVRAG/Fast-VRAG}.
\end{abstract}
\begin{figure*}[ht]
  \includegraphics[width=0.96\linewidth]{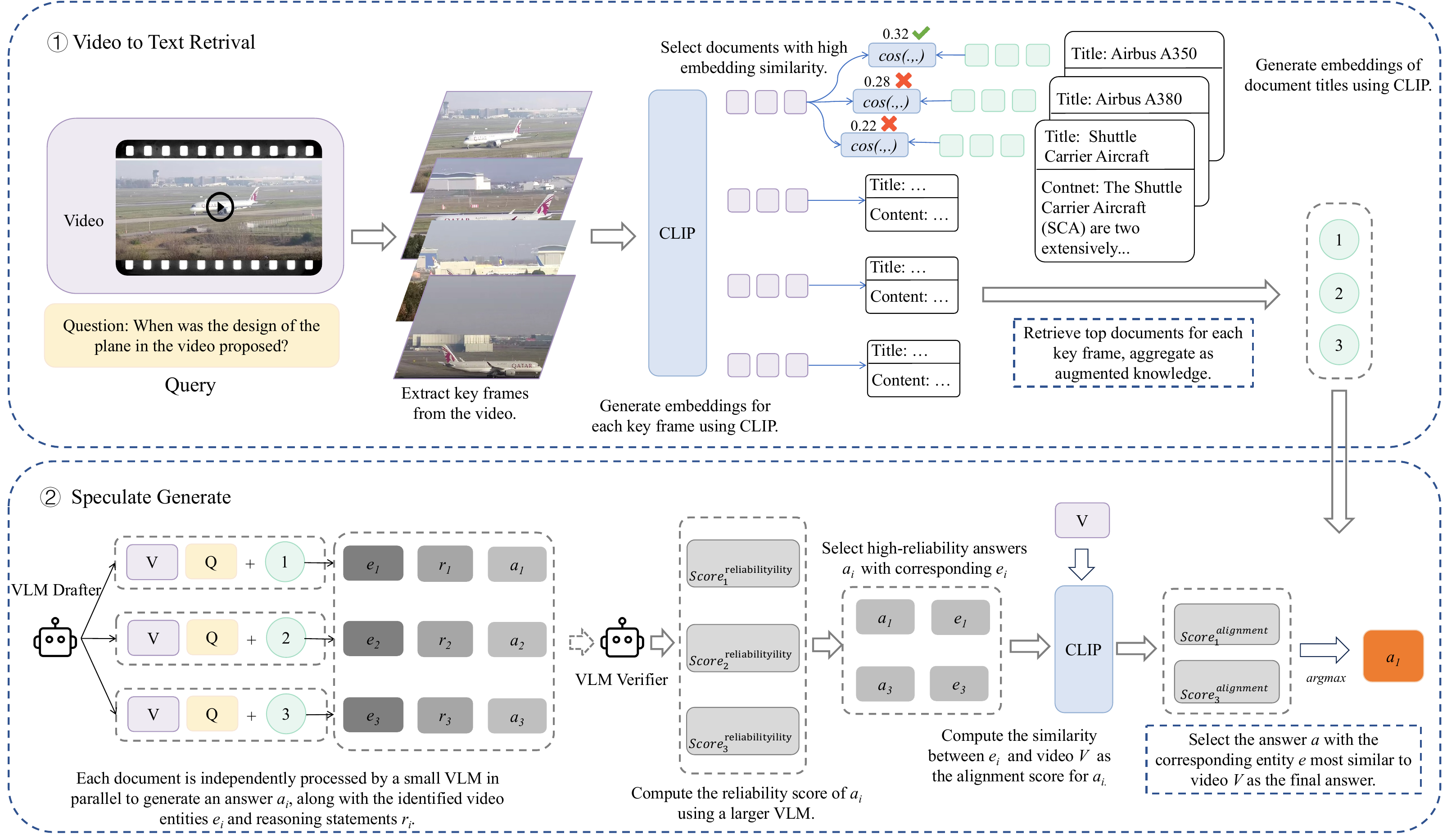} 

  \caption {Illustration of Video Speculate RAG.}
\end{figure*}
\section{Introduction}

Vision–Language Models~(VLMs) have advanced multimodal understanding, enabling machines to interpret visual inputs and generate contextually relevant text. Despite impressive progress~\cite{llava,minigpt,qwen2.5-VL,gpt4-o}, many real-world scenarios remain knowledge-intensive and require external information beyond the visual input.
Retrieval-Augmented Generation~(RAG) offers a promising solution, yet existing multimodal RAG approaches still incur high computational costs and struggle to maintain accuracy when integrating large knowledge sources or long-context inputs.
This motivates the need for an efficient generation strategy tailored to multimodal RAG.

To address such knowledge-intensive tasks, VLMs are increasingly combined with Retrieval-Augmented Generation~(RAG~\citealp{rag}), enabling them to retrieve relevant external information to support answer generation. In most existing approaches~\cite{fid,BlenderBot3,chen2022murag}, the retrieved passages are directly concatenated with the original multimodal inputs and processed jointly by the model. While this strategy enhances factual grounding and broadens the accessible knowledge scope, it also introduces several critical challenges. Traditional RAG pipelines often suffer from inefficiency, as inference latency and computational cost increase sharply with the number and length of retrieved documents. Moreover, in multimodal settings, the alignment between retrieved textual entities and visual entities is frequently imperfect, leading to mismatched reasoning and degraded accuracy. Overcoming these limitations is essential for building VLM–RAG systems that are both efficient and reliable, capable of reasoning coherently over complex visual and textual inputs.

We are motivated by speculative decoding~\cite{decoding—heads,wang2024speculative}, which is designed to improve efficiency by generating multiple candidate outputs in parallel using a lightweight draft model and then verifying or refining these candidates with a more accurate heavyweight model.
This separation of drafting and verification maintains high output quality while reducing inference latency compared with standard sequential decoding.
Our key idea is to let a lightweight model handle multimodal retrieval and answer generation, while invoking a stronger model only for calibration. When applied to VLM–RAG systems such as KVQA, this speculative decoding paradigm enables the draft stage to rapidly propose answers from retrieved documents, and the verification stage to refine and validate them, substantially reducing inference cost while preserving accuracy and visual grounding.

Moreover, we observe that naïve speculative decoding may still produce errors caused by fine-grained entity confusion, where superficially similar but incorrect entities in retrieved texts mislead the model. To address this, we introduce a fine-grained entity alignment mechanism: the drafter explicitly extracts entities and reasoning traces during answer generation, and the verifier measures alignment between candidate answers and video frames via CLIP-based similarity. This two-stage verification ensures that the final answer is both factually accurate and visually grounded, effectively mitigating entity-level errors while retaining the efficiency advantages of speculative decoding.

Finally, we evaluate VideoSpeculateRAG on two KVQA benchmarks, VideoSimpleQA~\cite{vqa} and Encyclopedic~VQA~\cite{evqa}, considering both answer accuracy and inference latency. Compared with the standard RAG framework, our approach achieves comparable or improved accuracy while providing nearly a $2\times$ speedup, highlighting its effectiveness in enhancing answer reliability and computational efficiency for complex multimodal reasoning tasks.

\section{Related Works}

\paragraph{Knowledge-aware VQA.}Knowledge-aware VQA refers to a class of question-answering tasks that require external knowledge, with questions and answers potentially spanning multiple modalities such as text, video, and images. \citet{okvqa,aokvqa,infoseek,evqa} provide sets of image-based questions whose answers cannot be directly inferred from the images themselves but instead rely on external textual knowledge. \citet{mmrag,m2rag} extend beyond textual knowledge: some of their questions require retrieving information from external images to answer. VideoSimpleQA~\cite{cao2025video} offers a collection of video–text question–answer pairs designed for evaluating question-answering systems in video-based contexts.

\paragraph{Speculative Decoding.}Speculative decoding~\cite{parallelDecoding,chen2023acceleratinglargelanguagemodel,leviathan2023fast} is an efficient generation strategy designed to accelerate large language model (LLM) inference by coupling a lightweight draft model with a more powerful target model. In conventional autoregressive decoding, the target model must generate tokens sequentially, which is computationally expensive and leads to high latency, especially for long sequences. Speculative decoding mitigates this inefficiency by allowing the draft model to rapidly produce multiple candidate tokens in parallel, effectively “speculating” on what the target model would generate. These proposed tokens are then passed to the target model for verification: if the target model’s predictions match the draft’s proposals, they are accepted directly, enabling the system to advance multiple tokens in a single step. If discrepancies occur, only the mismatched tokens are regenerated by the target model, ensuring correctness. This mechanism substantially reduces the number of costly forward passes through the large model while maintaining output fidelity, thereby offering a practical balance between computational efficiency and generation quality.

\FloatBarrier 
\begin{algorithm}[h]
\small
\caption{Video-to-Text Retrieval with Two-Stage Verification}
\label{alg1}

\KwIn{Video $V$, Question $Q$}
\KwOut{Final answer $a^\ast$}

$F \leftarrow \varnothing$ \tcp*{Initialize keyframe set}
\For{each frame $f_i$ in $V$}{
    \If{${sim}(f_i,f_{i-1}) < \theta$}{$F \leftarrow F \cup \{f_i\}$ \tcp*{Eq.(\ref{eq:frame_sample})}}
}

\For{each keyframe $f \in F$}{
    retrieve $T^\ast \leftarrow \text{Top-}K\ \text{sim}(f,t)$ \tcp*{Eq.(\ref{eq:clip_sim},\ref{eq:rag_docs})}
}

\For{each $t_i \in T^\ast$}{
    $e_i \leftarrow \text{Drafter}(V,  t_i)$ \tcp*{Eq.(\ref{eq:ett})}
    $r_i \leftarrow \text{Drafter}(V,Q, e_i, t_i)$ \tcp*{Eq.(\ref{eq:rationale})}
    $a_i \leftarrow \text{Drafter}(V,Q, e_i, r_i)$ \tcp*{Eq.(\ref{eq:answer})}
}

\For{each tuple $(a_i,r_i)$}{
    $\text{Score}_i^{\text{reliability}} \leftarrow \dfrac{p_i^{\text{Yes}}}{p_i^{\text{Yes}}+p_i^{\text{No}}}$ \tcp*{Eq.(\ref{eq:reliable})}
}

$A_H \leftarrow \{(a_i,e_i) \mid \text{Score}_i^{\text{reliability}} \ge \max_j \text{Score}_j^{\text{reliability}} - \delta\}$ \tcp*{Eq.(\ref{eq:AH})}

\For{each $a_i \in A_H$}{
    $\text{Score}_{\text{alignment}}(a_i) \leftarrow \max_{f\in F}\cos(\text{CLIP}(e_i),\text{CLIP}(f))$ \tcp*{Eq.(\ref{eq:vlm_rel},\ref{eq:clip_ett})}
}

$a^\ast \leftarrow \arg\max_{a_i \in A_H} \text{Score}_{\text{alignment}}(e_i)$ \tcp*{Select final answer}
\Return $a^\ast$

\end{algorithm}

\section{Methods}
In this work, we aim to enhance the efficiency and reliability of multimodal RAG by integrating speculative decoding with targeted answer verification. The core idea is to delegate most retrieval-conditioned generation to a lightweight VLM and invoke a stronger VLM only for calibration when needed. Section~\ref{sec-videorag} presents the proposed \emph{VideoSpeculateRAG} framework, and Section~\ref{sec-align} introduces an additional alignment enhancement to further improve answer accuracy.

\subsection{VideoSpeculateRAG}
\label{sec-videorag}

\paragraph{Multimodal Retrieval.}
Multimodal RAG aims to enhance video question answering by leveraging relevant external textual knowledge, enabling the model to reason over both visual content and supporting text. In this framework, the retrieval stage identifies and collects external documents that are semantically relevant to the video content and the posed question, providing the generation model with pertinent background knowledge. The subsequent generation stage produces the final answer by conditioning on both the video input and the retrieved textual knowledge, integrating multimodal information to generate accurate and contextually grounded responses.

To enable video-to-text retrieval, we first extract keyframes from the video. For each frame in video $V$, we compute its similarity with the previous frame. If the similarity falls below a predefined threshold, the frame is added to the keyframe set $F$, as shown in Eq.~(1):
\begin{equation}
  \label{eq:frame_sample}
    {sim}(f_i, f_{i-1}) < \theta ,  F \leftarrow F \cup \{f_i\}.
\end{equation}
Here, histograms are used to represent inter-frame similarity.
For each keyframe $f$, we retrieve its top-$k$ most similar texts and aggregate them into a text set $T$. Given an image frame and a text, their similarity is computed using CLIP embeddings and cosine similarity, as shown in Eq.~(2):
\begin{equation}
  \label{eq:clip_sim}
    {sim}(f, t) = \cos\!\big( \text{CLIP}(f), \text{CLIP}(t) \big).
\end{equation}
We select the top-$K$ texts from the set $T$ with the highest similarity scores, which serve as external knowledge to enhance the VLM input, as shown in Eq.~(3):
\begin{equation}
  \label{eq:rag_docs}
    T^\ast = \underset{t \in T}{\text{Top-}K}\; \text{sim}(f, t).
\end{equation}

Finally, the retrieved external texts are concatenated after the question, and the VLM generates the answer $A$ based on the video $V$, the question $Q$, and the text set $T^*$:
\begin{equation}
a \sim \text{VLM}(Q, V, T^*).
\end{equation}
\paragraph{Speculate Generation.}
Speculative decoding extends the generation stage of standard multimodal RAG by introducing a two-stage verification process that improves both efficiency and answer reliability. In this framework, a lightweight model first rapidly generates draft answers for each retrieved document, producing multiple candidate responses in parallel. These drafts are then evaluated during the verification stage, where a scoring mechanism determines the most reliable answer to be selected as the final output.

Unlike standard multimodal RAG, after retrieving the external document set $T^*$, we do not concatenate all documents into a single input. Instead, each document $t_i \in T^*$ is appended to a separate input, and these inputs are processed in parallel by a lightweight draft VLM to rapidly generate a set of draft answers:
\begin{equation}
   A_{\text{draft}} = \{a_i \sim \text{VLM}_{\text{Drafter}}(Q, V, t_i) \mid t_i \in T^* \}.
\end{equation}
Next, a larger verifier VLM evaluates each draft answer $a_i \in \mathcal{A}_{\text{draft}}$, computing a corresponding score:
\begin{equation}
    \text{score}_i = \{\text{VLM}_{\text{Verifier}}(Q, V, a_i) \mid a_i \in A_{\text{draft}}\}.
\end{equation}
Finally, the draft with the highest score is selected as the final answer:
\begin{equation}
a^* = \text{argmax}(\{ \text{score}_i \mid a_i \in A_{\text{draft}} \}).
\end{equation}
By parallelizing draft generation over individual documents, the framework mitigates the computational overhead and memory burden induced by long-context concatenation, thereby enhancing efficiency while preserving answer reliability.

\begin{figure*}[t]
  \includegraphics[width=1.0\linewidth]{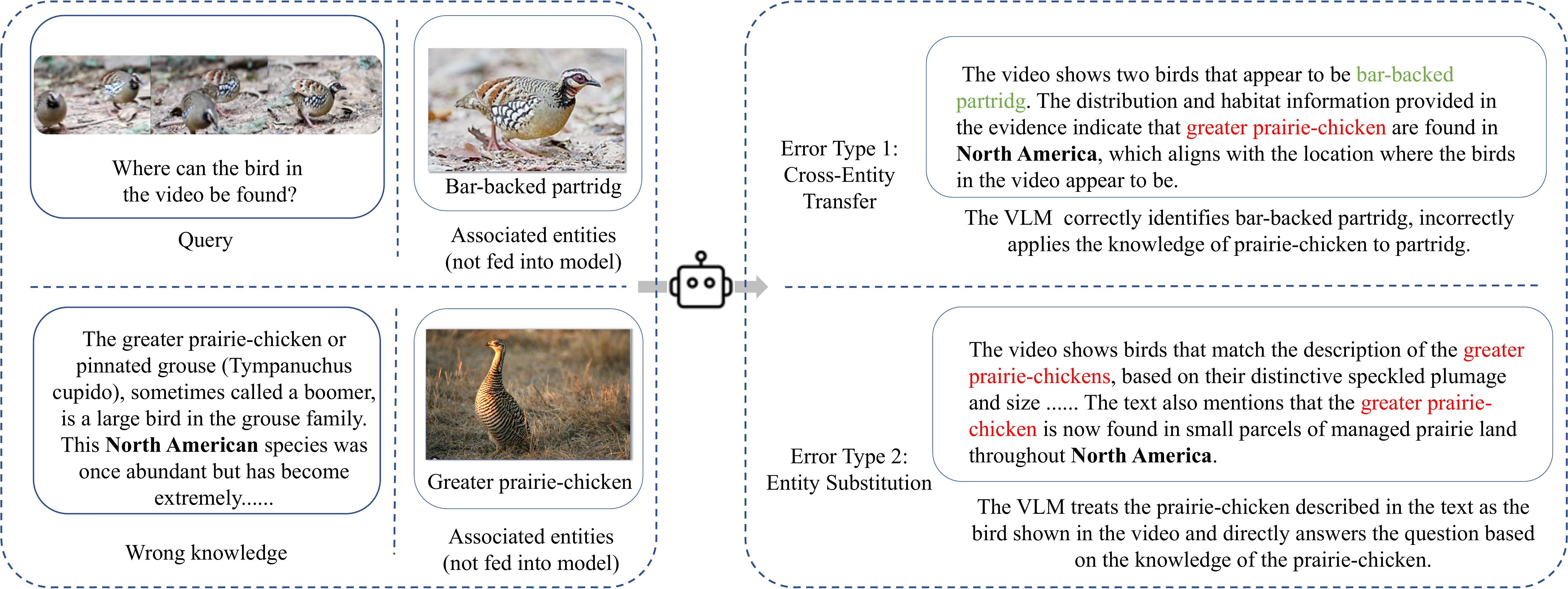}
  \caption{Error Analysis. \textcolor{red}{RED} color stands for entities inconsistent with the video content, while \textcolor{green}{GREEN} color refers entities consistent with the video.}
  \label{dataset}
\end{figure*}

\begin{figure*}[ht]
  \includegraphics[width=0.96\linewidth]{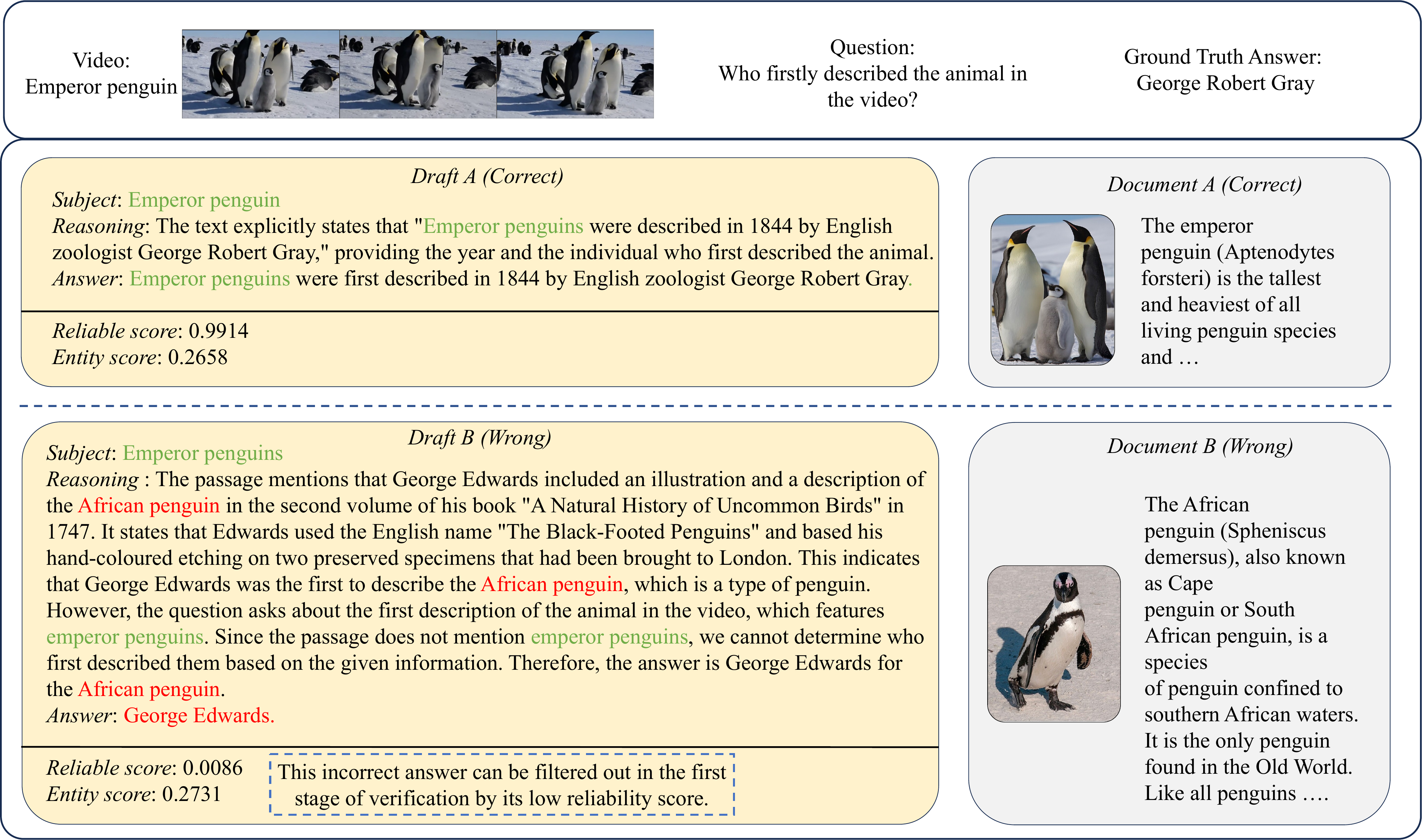}
  \caption{An illustrative example of how our method detects Cross-Entity Transfer.}
  \label{image:partial}
\end{figure*}

\begin{figure*}[t]
\centering
  \includegraphics[width=0.96\linewidth]{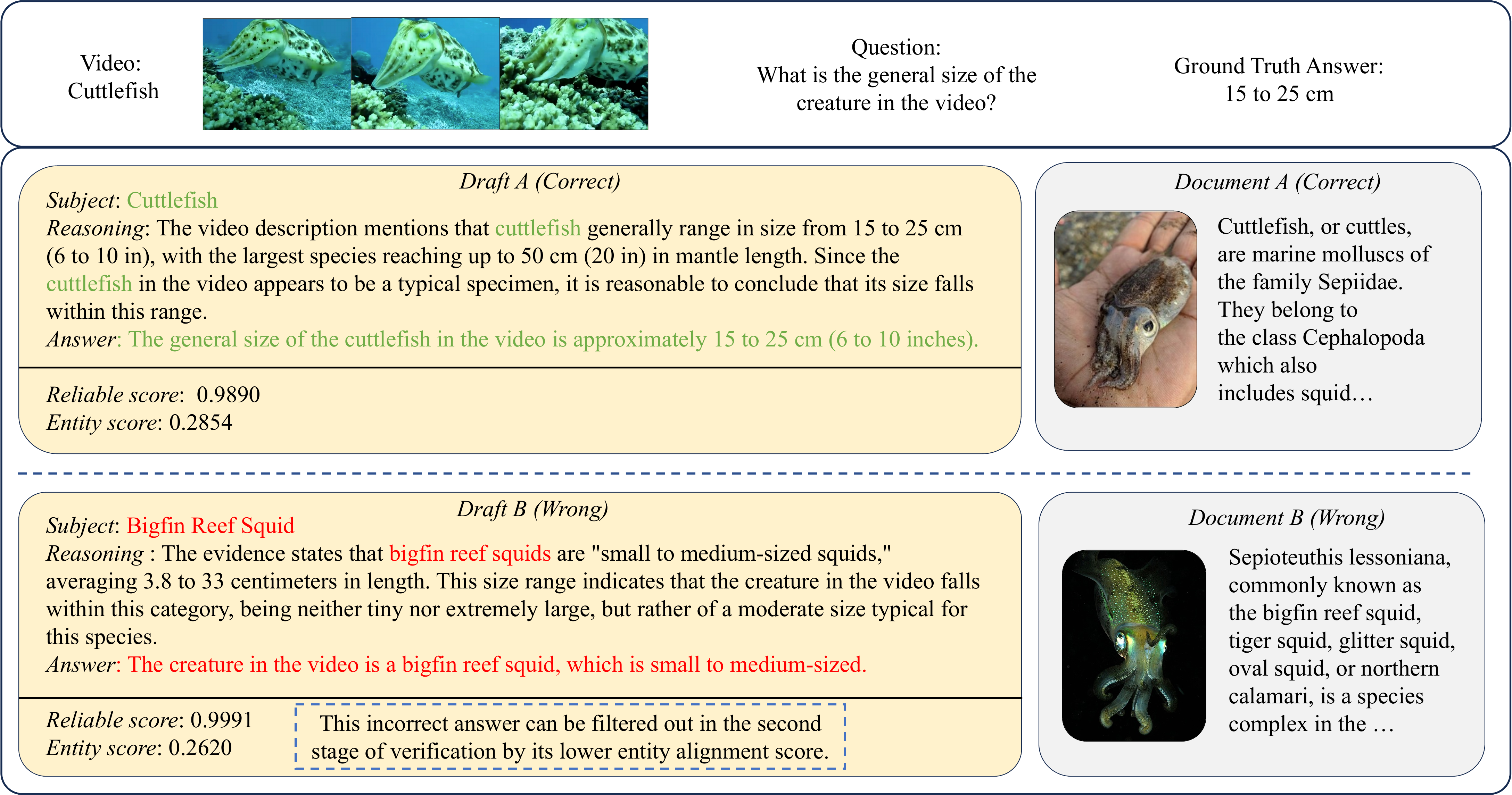}
  \caption{An illustrative example of how our method detects Entity Substitution.}
  \label{image:full}
\end{figure*}

\subsection{{Fine-grained Entity Alignment}}
\label{sec-align}

While speculative decoding efficiently generates multiple draft answers and allows the verifier to select the most likely one, this process alone may not fully address errors arising from fine-grained entity confusion. Specifically, some drafts can appear plausible yet contain subtle mistakes caused by retrieved documents referencing entities that are visually or semantically similar but not identical to those in the video. To systematically detect such errors during verification, we introduce a fine-grained entity alignment mechanism.

\paragraph{Error Analysis.}
Given a video $V$, suppose that an erroneous document $T_\text{error}$ containing entities that are superficially similar to but inconsistent with the actual video content is provided to the VLM. In this situation, the model can be misled into producing incorrect answers. We categorize such error scenarios into two representative types, assuming that the actual entity in the video is $E_\text{actual}$ and the erroneous entity in the text is $E_\text{error}$:

$\bullet$ \emph{Cross-Entity Transfer}: The VLM successfully identifies the actual visual entity $E_\text{actual}$ within the video, yet its reasoning is influenced by external knowledge associated with $E_\text{error}$. This contamination leads the model to apply attributes or contextual information from the erroneous entity to the correct one. As a result, the generated answer often carries subtle traces of $E_\text{error}$, giving rise to inconsistencies at the entity level. While such answers may retain a degree of surface-level alignment with the video, they ultimately reflect a fragile and unreliable grounding.

$\bullet$ \emph{Entity Substitution}: The VLM is fully influenced by the erroneous document, resulting in a final answer that is grounded entirely on the incorrect entity $E_\text{error}$ rather than the true visual entity $E_\text{actual}$. The reasoning process is dominated by the misleading text, producing outputs that are internally coherent and linguistically fluent, yet detached from the actual content of the video. The answer may superficially appear plausible, but it fails to reflect the correct fine-grained entities and their relationships present in the visual input. As a consequence, while the response may seem consistent and self-contained, it systematically replaces the authentic entity information with spurious attributes. 

\paragraph{Structured Draft Reasoning.}
To enable the verification stage to identify such errors, we design a multi-hop reasoning procedure in which the draft VLM explicitly extracts entities from the video and structures its reasoning trajectory. This structured information equips the subsequent verification with the necessary signals to detect misleading drafts. Given a video $v$, a question $Q$, and a retrieved text $t_i$, the drafter first extracts the salient entity $e_i$ from the video:
\begin{equation}
\label{eq:ett}
e_i = \text{VLM}_\text{Drafter}(V, t_i).
\end{equation}
Conditioned on this entity and the question, the model incorporates the retrieved text $t_i$ as evidence to form an intermediate reasoning statement $r_i$, explicitly linking the entity, the question, and the supporting text:
\begin{equation}
\label{eq:rationale}
r_i = \text{VLM}_\text{Drafter}(V,Q, e_i, t_i).
\end{equation}
Finally, leveraging both the extracted entity and the reasoning statement, the drafter produces a candidate answer $a_i$:
\begin{equation}
\label{eq:answer}
a_i = \text{VLM}_\text{Drafter}(V,Q, e_i, r_i),  A_\text{Draft} \gets A_\text{Draft} \cup \{a_i\}.
\end{equation}

\paragraph{Two-Stage Verification.}
After generating candidate answers $a_i$ along with the corresponding evidence $e_i$ and reasoning statements $r_i$, these drafts are evaluated through a two-stage verification process to select the final answer. In the first stage, the verifier VLM assesses the reliability of each candidate, determining whether the reasoning supports the answer. A single forward pass computes the probability that the first generated token is ``Yes'':
\begin{equation}
  \label{eq:yes}
    p_i^\text{Yes} = P(\text{``Yes''} \mid Q,V, a_i, e_i, r_i),
\end{equation}
and similarly for ``No''. The reliability score is then calculated as:
\begin{equation}
  \label{eq:reliable}
    \text{Score}_i^\text{reliable} = \frac{p_i^\text{Yes}}{p_i^\text{Yes} + p_i^\text{No}}.
\end{equation}
Then we select candidate answers $A_H$ whose scores satisfy the following requirement:
\begin{equation}
  \label{eq:AH}
\text{Score}_i^\text{reliable}\ge \max_{a_j \in A_\text{Draft}} (\text{Score}_j^\text{reliable}(a_j)) - \delta,
\end{equation}
where $\delta$ denotes a tolerance margin from the maximum reliability score.

Even within $A_H$, the entities described in the candidate answers may still differ subtly from the actual video entities. To mitigate this discrepancy, we compute an entity alignment score for each candidate answer $a_i \in A_H$ by measuring the similarity between its corresponding entity $e_i$ and the associated video frames $f_i$ using CLIP:
\begin{equation}
  \label{eq:vlm_rel}
    {sim}(e_i, f_j) = \cos\!\big( \text{CLIP}(e_i), \text{CLIP}(f_j) \big).
\end{equation}
The maximum similarity over all frames is taken as the alignment score:
\begin{equation}
  \label{eq:clip_ett}
   \text{Score}_\text{alignment}(a_i) = \max_{f_i \in F} \text{sim}(e_i, f_i).
\end{equation}
The candidate with the highest alignment score is then selected as the final answer, as illustrated in Algorithm~\ref{alg1}.

\section{Experiments}
We conduct experiments on video and image datasets to evaluate the accuracy and reasoning speed of our system on the KVQA task.
\subsection{Datasets}
\textbf{VideoSimpleQA.} VideoSimpleQA~\cite{cao2025video} is a video-based KVQA dataset that consists of videos collected from Wikimedia Commons and questions extracted from corresponding Wikipedia entries. For our experiments, we curate a subset of VideoSimpleQA by selecting videos that are clear and contain unambiguous entities, along with their associated questions, to serve as our video KVQA dataset.
\\
\textbf{Encyclopedic VQA.} Due to the lack of established benchmarks for video-based KVQA, we additionally adopt an image-based KVQA dataset, Encyclopedic VQA~\cite{evqa}, hereafter referred to as EncVQA. This dataset is constructed from images in the iNaturalist 2021~\cite{van2021benchmarking} and Google Landmarks Dataset V2 collections~\cite{weyand2020google}, with questions derived from corresponding Wikipedia entries. 

\subsection{Baselines}

\paragraph{No RAG.}
The video and the question are directly fed into the VLM, which produces the answer solely based on its internal knowledge. This baseline evaluates the inherent capability of VLMs to answer knowledge-intensive questions without external documents. We evaluate several open-source VLMs, including \emph{Qwen2.5-VL-Instruct 3B}, \emph{Qwen2.5-VL-Instruct 32B}, \emph{LLaVA-NeXT-Video 34B}, and \emph{InternVL3 38B}. More implementation details can be found in Appendix~\ref{app-model}.

\paragraph{Standard RAG.} Retrieved documents are directly concatenated into the prompt alongside the visual content and the question, and then passed into the VLM to produce answers. This baseline assesses the model’s ability to comprehend the retrieved text and align it with the video content. The same set of open-source VLMs used in the No RAG baseline are evaluated here.

\subsection{Implementation Details}
For each dataset, we employ \emph{clip-vit-large-patch14-336} in the retrieval stage to compute the similarity between the text and visual content, retrieving $k=3$ documents for each query. We employ \emph{Qwen2.5-VL-Instruct-3B} as the Drafter and \emph{Qwen2.5-VL-Instruct-32B} as the Verifier, with $\delta$ set to $0.05$, which yields the best empirical performance in our experiments.

\subsection{Main Results}

\begin{table*}
  \centering
  \small
  \resizebox{\linewidth}{!}{
  \begin{tabular}{lcccccc}
    \toprule
    \textbf{Methods}           & \textbf{VideoSimpleQA} & \textbf{Latency (s)} & \textbf{EncVQA} & \textbf{Latency (s)}&\textbf{Avg Accuracy}&\textbf{Avg Latency (s)}\\
    \midrule
    \rowcolor{gray!10}\multicolumn{7}{c}{\it \textbf{No RAG}}\\ \midrule
    $\text{Qwen-2.5VL-Instruct-3B}$  & 37.62 &2.15 &14.00 &3.30&25.81&2.72\\
    $\text{Qwen-2.5VL-Instruct-32B}$&57.73&19.18&22.00&20.13&39.87&19.65\\
    $\text{LLaVA-NeXT-Video-34B}$&27.31&29.24&8.50&25.68&17.92&27.46 \\
    $\text{InternVL-3-38B}$&40.20&4.06&13.5&5.61&26.85&4.84 \\
    \midrule
    \rowcolor{gray!10}\multicolumn{7}{c}{\it \textbf{Standard RAG}}\\ \midrule
    $\text{Qwen-2.5VL-Instruct-3B}$& 75.26&4.84 &39.00&4.76&57.13&4.80\\
    $\text{Qwen-2.5VL-Instruct-32B}$&\textbf{91.30}&47.72&\underline{46.75}&40.42&\underline{69.03}&44.07\\
    $\text{LLaVA-NeXT-Video-34B}$&76.80&56.54&37.50&51.37&57.15&53.96\\
    $\text{InternVL-3-38B}$&78.86&14.00&43.50&7.75&61.18&10.88\\
    
    \midrule
    \rowcolor{gray!10}\multicolumn{7}{c}{\it \textbf{Speculate RAG}}\\ \midrule
    $\text{Qwen-2.5VL-Instruct-3B+7B}$&82.60&23.04&41.00&13.03&61.80&19.54\\
    $\text{Qwen-2.5VL-Instruct-3B+32B}$&\underline{91.12}&25.74&\textbf{47.64}&16.61&\textbf{69.38}&21.18 \\
    \bottomrule
  \end{tabular}}
  \caption{
    Main results of vision-language models on knowledge-intensive video QA tasks. The annotation after each model indicates the model size. Values under each task indicate accuracy~(higher is better), while the latency represents inference time~(lower is better). The best results are highlighted in \textbf{bold} while the second best using \underline{underline}.
  }
  \label{table:main}
\end{table*}
\paragraph{VideoSpeculateRAG can achieve high accuracy on KVQA tasks.}
As shown in Table~\ref{table:main}, without external knowledge, VLMs struggle to answer questions in both VideoSimpleQA and EncVQA: for instance, \textit{Qwen2.5-VL-Instruct-32B} achieves only 57.73\% and 22.00\% accuracy on the two datasets, respectively. With external knowledge augmentation, performance improves substantially. On VideoSimpleQA and EncVQA, \textit{Qwen2.5-VL-Instruct-32B (RAG)} achieves 91.30\% and 46.50\% accuracy, corresponding to relative improvements of 58.15\% and 111.36\% over the non-RAG setting. 
Under the speculative paradigm, our method matches or even surpasses the performance of standard RAG. Specifically, \textit{Qwen2.5-VL-Instruct-32B (VideoSpeculateRAG)} attains 91.12\% accuracy on VideoSimpleQA—comparable to the 91.30\% of standard RAG—and 47.64\% accuracy on EncVQA, outperforming the 46.75\% of standard RAG.

\paragraph{VideoSpeculateRAG can effectively reduce the inference latency of RAG.}Our experiments show that VLM inference time increases substantially with longer input contexts. After augmenting with external documents, \textit{Qwen2.5-VL-Instruct-32B} exhibits a latency increase from 19.18s to 47.72s on VideoSimpleQA, and from 20.13s to 40.42s on EncVQA. While, \textit{Qwen2.5-VL-Instruct-3+32B (VideoSpeculateRAG)} significantly alleviates this latency growth, achieving 25.74s and 16.61s on the two datasets, corresponding to reductions of 46.06\% and 58.90\% compared to \textit{Qwen-32B (RAG)}, and even outperforming the non-RAG in terms of speed. This improvement primarily stems from two factors: (i) the lightweight draft model processes inputs efficiently, and (ii) splitting the retrieved documents mitigates the effect of input context expansion.

\subsection{Ablation Study}
\begin{table}[h]
\centering
\small
\resizebox{\linewidth}{!}{
\begin{tabular}{lcccc}
\toprule
\textbf{Methods} & {\textbf{VideoSimpleQA}} & \textbf{Latency} & \textbf{EncVQA} & \textbf{Latency} \\
\midrule
Ours              & 91.12 & 25.74 & 47.64 & 16.61 \\
w/o ett       & 85.30 & 24.14 & 38.02 & 15.31 \\
w/o rel     & 65.71 & 21.62 & 43.22 & 12.46 \\
Random            & 47.59 & 20.85 & 27.08 & 11.59 \\
\bottomrule
\end{tabular}}
\caption{Ablation studies. Where ``ett'' stands for Entity Alignment Score while ``rel'' stands for reliability score.}
\end{table}
In the ablation study, we progressively remove the reliability score and the entity alignment score. Removing the reliability score causes a sharp accuracy drop (25.41\% on VideoSimpleQA and 4.44\% on EncVQA), with a latency reduction of about 4s, mainly due to skipping one VLM forward pass. Removing the alignment score decreases accuracy by 5.82\% and 9.62\% on the two datasets, while reducing latency by about 1s, since CLIP is faster than the 32B VLM. When both components are removed, the accuracy of \emph{VideoSpeculateRAG} collapses to roughly half of the original. These results confirm that reliability estimation and entity alignment are both critical and complementary.

\subsection{Further Analysis}
\paragraph{Case Studies.}

We illustrate how our framework addresses the errors analyzed in Section~\ref{sec-align} through two representative examples.
In Figure~\ref{image:partial}, the video shows an emperor penguin. Retrieved documents cover both emperor and African penguins. Draft A and Draft B achieve similar entity alignment, but Draft B incorporates incorrect external knowledge and thus receives a much lower ~\textbf{reliability score}. Leveraging this score, our framework filters out Draft B in the first-stage verification and selects Draft A as the final answer.
In Figure~\ref{image:full}, the video shows a cuttlefish, but documents about both cuttlefish and squid are retrieved. Draft A correctly recognizes the cuttlefish, while Draft B is fully misled and confidently predicts squid. Both pass the first-stage reliability check, but in the second stage Draft A achieves a higher ~\textbf{entity alignment score}, allowing our framework to select it as the final answer.
These results further highlight the necessity of the proposed two-stage verification.

\paragraph{Analysis of Scoring Mechanisms.}
\begin{table}[h]
\centering
\small
\resizebox{\linewidth}{!}{
\begin{tabular}{lccc}
\toprule
\textbf{Strategies} & \textbf{Model} & {\textbf{VideoSimpleQA}}  & \textbf{EncVQA}  \\
\midrule
    Self Cossistent &3B+7B&65.59&29.00 \\
    &3B+32B&68.63&36.50 \\ \midrule
    Addition&3B+7B&73.91&42.00\\
    &3B+32B&87.57&46.39\\ \midrule
    Invert&3B+7B&73.33&40.10\\
    &3B+32B&74.86&41.23\\ 
\bottomrule
\end{tabular}}
\caption{Comparison of verification scoring strategies. We compare three key scoring methods include ``Self consistent'', ``Addition'' and ``Invert''.}
\label{table:strategy}
\end{table}
To study the impact of verification strategies, we compare three scoring mechanisms using different model setups, where ``3B+7B'' and ``3B+32B'' denote a 3B draft model paired with a 7B or 32B verifier.
The \textit{self-consistent} strategy, adopted from Speculative RAG, relies solely on the verifier’s internal likelihoods. The \textit{addition} strategy directly sums the reliability and alignment scores, while the \textit{invert} strategy first filters candidates by alignment score and then ranks them by reliability. As shown in Table~\ref{table:strategy}, the \textit{self-consistent} method performs the worst, as its confidence measure is affected by token length and fails to capture fine-grained multimodal alignment. The \textit{addition} strategy achieves the best overall results, though imperfect calibration may arise since reliability and alignment are measured in different semantic spaces. The \textit{invert} approach yields moderate but unstable gains due to the limited discriminative power of the entity alignment score in the first-stage filtering.
\paragraph{Analysis of settings of $\delta$.}
\begin{figure}[t]
  \includegraphics[width=\columnwidth]{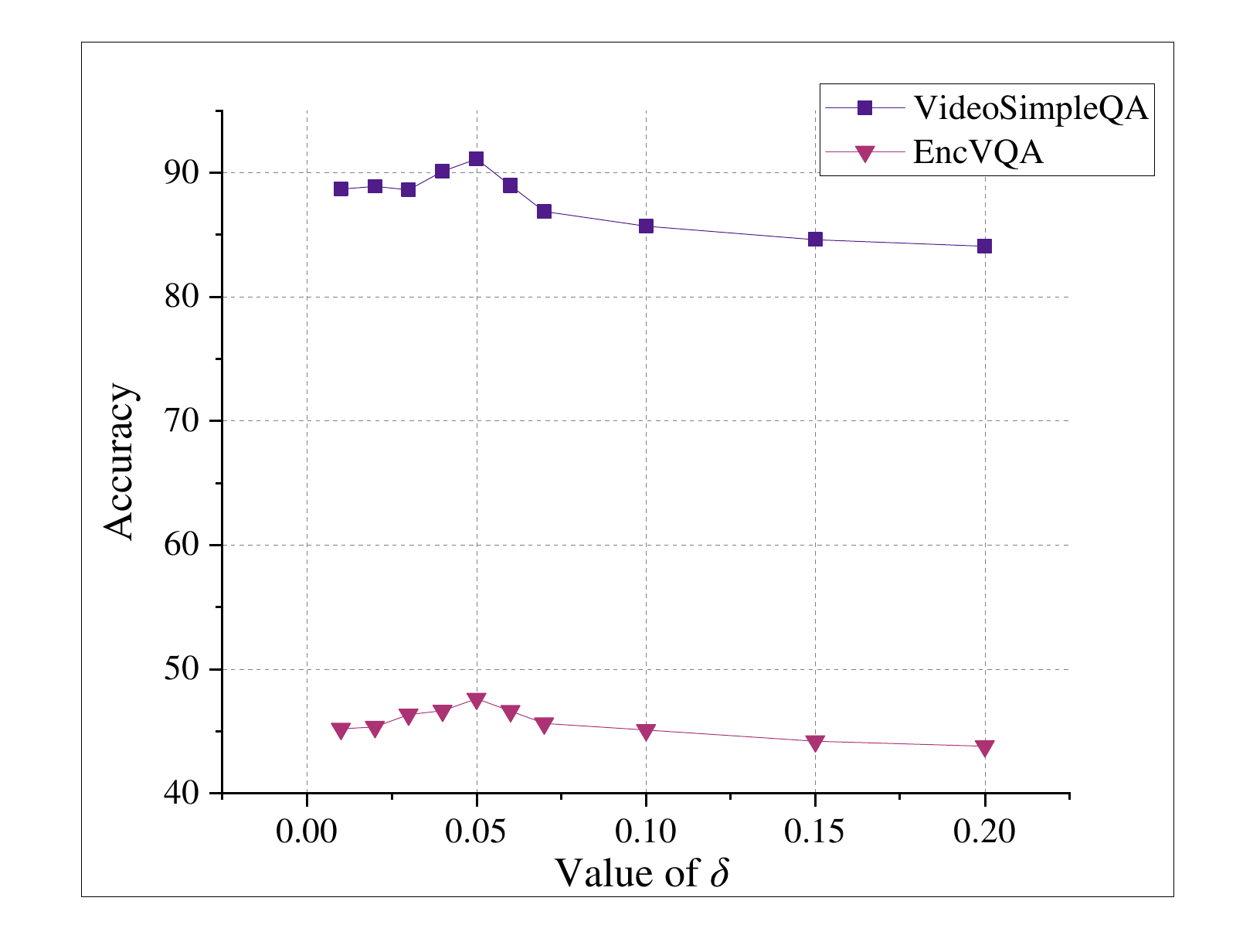}
  \caption{Accuracy variation with different $\delta$ values}
  \label{fig:delta}
\end{figure}
The tolerance margin $\delta$ in Eq.~\ref{eq:AH} controls the threshold for selecting high-reliability candidates. We evaluate the impact of different $\delta$ values on answer accuracy to analyze how this parameter affects the performance of \textit{VideoSpeculateRAG}. As shown in Table~\ref{fig:delta}, the accuracy on both datasets first increases and then decreases as $\delta$ grows, reaching its peak when $\delta = 0.05$. Smaller $\delta$ values cause the system to rely almost exclusively on reliability scores, leading to more frequent \textit{Entity Substitution} errors, whereas larger $\delta$ values make the system depend excessively on entity alignment, resulting in more \textit{Cross-Entity Transfer} errors.

\section{Conclusion}
In this work, we propose VideoSpeculateRAG, a multimodal RAG framework that combines speculative decoding with two-stage answer verification, where a lightweight VLM drafts retrieval-conditioned answers and a stronger VLM calibrates them to reduce inference latency without sacrificing accuracy.
To further enhance reliability, we introduce an entity-aware verification strategy to resolve fine-grained entity confusion. Experiments on VideoSimpleQA and Encyclopedic VQA show that our method outperforms standard RAG and alternative speculative methods, achieving higher accuracy with substantially lower inference cost. We believe this approach paves the way for real-time, knowledge-intensive video interaction in future multimodal systems.

\section*{Limitations}
For a fair comparison with prior work, we evaluate VideoSpeculateRAG using widely adopted foundation vision–language models. However, the performance of video RAG may be constrained by the underlying model capability, which we leave for future investigation. In addition, our experiments focus on knowledge-intensive video QA datasets, while real-world scenarios can be more diverse and complex; extending the analysis to broader tasks and settings remains an important direction for future work.

\bibliography{custom}

@InProceedings{aokvqa,
    author="Schwenk, Dustin
    and Khandelwal, Apoorv
    and Clark, Christopher
    and Marino, Kenneth
    and Mottaghi, Roozbeh",
    editor="Avidan, Shai
    and Brostow, Gabriel
    and Ciss{\'e}, Moustapha
    and Farinella, Giovanni Maria
    and Hassner, Tal",
    title="A-OKVQA: A Benchmark for Visual Question Answering Using World Knowledge",
    booktitle="Computer Vision -- ECCV 2022",
    year="2022",
    publisher="Springer Nature Switzerland",
    address="Cham",
    pages="146--162",
    isbn="978-3-031-20074-8"
}

@inproceedings{okvqa,
  title={Ok-vqa: A visual question answering benchmark requiring external knowledge},
  author={Marino, Kenneth and Rastegari, Mohammad and Farhadi, Ali and Mottaghi, Roozbeh},
  booktitle={Proceedings of the IEEE/cvf conference on computer vision and pattern recognition},
  pages={3195--3204},
  year={2019}
}

@misc{infoseek,
      title={Can Pre-trained Vision and Language Models Answer Visual Information-Seeking Questions?}, 
      author={Yang Chen and Hexiang Hu and Yi Luan and Haitian Sun and Soravit Changpinyo and Alan Ritter and Ming-Wei Chang},
      year={2023},
      eprint={2302.11713},
      archivePrefix={arXiv},
      primaryClass={cs.CV},
      url={https://arxiv.org/abs/2302.11713}, 
}

@article{mmrag,
    author = {Zhan, Zaifu and Wang, Jun and Zhou, Shuang and Deng, Jiawen and Zhang, Rui},
    title = {MMRAG: multi-mode retrieval-augmented generation with large language models for biomedical in-context learning},
    journal = {Journal of the American Medical Informatics Association},
    pages = {ocaf128},
    year = {2025},
    month = {08},
    issn = {1527-974X},
    doi = {10.1093/jamia/ocaf128},
    url = {https://doi.org/10.1093/jamia/ocaf128},
    eprint = {https://academic.oup.com/jamia/advance-article-pdf/doi/10.1093/jamia/ocaf128/63948576/ocaf128.pdf},
}

@inproceedings{evqa,
  title={Encyclopedic vqa: Visual questions about detailed properties of fine-grained categories},
  author={Mensink, Thomas and Uijlings, Jasper and Castrejon, Lluis and Goel, Arushi and Cadar, Felipe and Zhou, Howard and Sha, Fei and Araujo, Andr{\'e} and Ferrari, Vittorio},
  booktitle={Proceedings of the IEEE/CVF International Conference on Computer Vision},
  pages={3113--3124},
  year={2023}
}

@article{m2rag,
  title={Multi-modal Retrieval Augmented Multi-modal Generation: Datasets, Evaluation Metrics and Strong Baselines},
  author={Ma, Zi-Ao and Lan, Tian and Tu, Rong-Cheng and Hu, Yong and Zhu, Yu-Shi and Zhang, Tong and Huang, Heyan and Wu, Zhijing and Mao, Xian-Ling},
  journal={arXiv preprint arXiv:2411.16365},
  year={2024}
}

@article{cao2025video,
  title={Video simpleqa: Towards factuality evaluation in large video language models},
  author={Cao, Meng and Hu, Pengfei and Wang, Yingyao and Gu, Jihao and Tang, Haoran and Zhao, Haoze and Dong, Jiahua and Yu, Wangbo and Zhang, Ge and Reid, Ian and others},
  journal={arXiv preprint arXiv:2503.18923},
  year={2025}
}

@inproceedings{parallelDecoding,
 author = {Stern, Mitchell and Shazeer, Noam and Uszkoreit, Jakob},
 booktitle = {Advances in Neural Information Processing Systems},
 editor = {S. Bengio and H. Wallach and H. Larochelle and K. Grauman and N. Cesa-Bianchi and R. Garnett},
 pages = {},
 publisher = {Curran Associates, Inc.},
 title = {Blockwise Parallel Decoding for Deep Autoregressive Models},
 url = {https://proceedings.neurips.cc/paper_files/paper/2018/file/c4127b9194fe8562c64dc0f5bf2c93bc-Paper.pdf},
 volume = {31},
 year = {2018}
}

@misc{chen2023acceleratinglargelanguagemodel,
      title={Accelerating Large Language Model Decoding with Speculative Sampling}, 
      author={Charlie Chen and Sebastian Borgeaud and Geoffrey Irving and Jean-Baptiste Lespiau and Laurent Sifre and John Jumper},
      year={2023},
      eprint={2302.01318},
      archivePrefix={arXiv},
      primaryClass={cs.CL},
      url={https://arxiv.org/abs/2302.01318}, 
}

@inproceedings{leviathan2023fast,
  title={Fast inference from transformers via speculative decoding},
  author={Leviathan, Yaniv and Kalman, Matan and Matias, Yossi},
  booktitle={International Conference on Machine Learning},
  pages={19274--19286},
  year={2023},
  organization={PMLR}
}

@article{wang2024speculative,
  title={Speculative rag: Enhancing retrieval augmented generation through drafting},
  author={Wang, Zilong and Wang, Zifeng and Le, Long and Zheng, Huaixiu Steven and Mishra, Swaroop and Perot, Vincent and Zhang, Yuwei and Mattapalli, Anush and Taly, Ankur and Shang, Jingbo and others},
  journal={arXiv preprint arXiv:2407.08223},
  year={2024}
}

@article{minigpt,
  title={Minigpt-4: Enhancing vision-language understanding with advanced large language models},
  author={Zhu, Deyao and Chen, Jun and Shen, Xiaoqian and Li, Xiang and Elhoseiny, Mohamed},
  journal={arXiv preprint arXiv:2304.10592},
  year={2023}
}

@article{llava,
  title={Video-llava: Learning united visual representation by alignment before projection},
  author={Lin, Bin and Ye, Yang and Zhu, Bin and Cui, Jiaxi and Ning, Munan and Jin, Peng and Yuan, Li},
  journal={arXiv preprint arXiv:2311.10122},
  year={2023}
}

@inproceedings{clip,
  title={Learning transferable visual models from natural language supervision},
  author={Radford, Alec and Kim, Jong Wook and Hallacy, Chris and Ramesh, Aditya and Goh, Gabriel and Agarwal, Sandhini and Sastry, Girish and Askell, Amanda and Mishkin, Pamela and Clark, Jack and others},
  booktitle={International conference on machine learning},
  pages={8748--8763},
  year={2021},
  organization={PmLR}
}

@misc{qwen2.5-VL,
    title = {Qwen2.5-VL},
    url = {https://qwenlm.github.io/blog/qwen2.5-vl/},
    author = {Qwen Team},
    month = {January},
    year = {2025}
}

@article{gpt4-o,
  title={Gpt-4o system card},
  author={Hurst, Aaron and Lerer, Adam and Goucher, Adam P and Perelman, Adam and Ramesh, Aditya and Clark, Aidan and Ostrow, AJ and Welihinda, Akila and Hayes, Alan and Radford, Alec and others},
  journal={arXiv preprint arXiv:2410.21276},
  year={2024}
}

@inproceedings{vqa,
  title={Vqa: Visual question answering},
  author={Antol, Stanislaw and Agrawal, Aishwarya and Lu, Jiasen and Mitchell, Margaret and Batra, Dhruv and Zitnick, C Lawrence and Parikh, Devi},
  booktitle={Proceedings of the IEEE international conference on computer vision},
  pages={2425--2433},
  year={2015}
}

@article{rag,
  title={Retrieval-augmented generation for knowledge-intensive nlp tasks},
  author={Lewis, Patrick and Perez, Ethan and Piktus, Aleksandra and Petroni, Fabio and Karpukhin, Vladimir and Goyal, Naman and K{\"u}ttler, Heinrich and Lewis, Mike and Yih, Wen-tau and Rockt{\"a}schel, Tim and others},
  journal={Advances in neural information processing systems},
  volume={33},
  pages={9459--9474},
  year={2020}
}

@article{decoding—heads,
  title={Medusa: Simple llm inference acceleration framework with multiple decoding heads},
  author={Cai, Tianle and Li, Yuhong and Geng, Zhengyang and Peng, Hongwu and Lee, Jason D and Chen, Deming and Dao, Tri},
  journal={arXiv preprint arXiv:2401.10774},
  year={2024}
}

@article{fid,
  title={Leveraging passage retrieval with generative models for open domain question answering},
  author={Izacard, Gautier and Grave, Edouard},
  journal={arXiv preprint arXiv:2007.01282},
  year={2020}
}

@article{BlenderBot3,
  title={Blenderbot 3: a deployed conversational agent that continually learns to responsibly engage},
  author={Shuster, Kurt and Xu, Jing and Komeili, Mojtaba and Ju, Da and Smith, Eric Michael and Roller, Stephen and Ung, Megan and Chen, Moya and Arora, Kushal and Lane, Joshua and others},
  journal={arXiv preprint arXiv:2208.03188},
  year={2022}
}

@article{chen2022murag,
  title={Murag: Multimodal retrieval-augmented generator for open question answering over images and text},
  author={Chen, Wenhu and Hu, Hexiang and Chen, Xi and Verga, Pat and Cohen, William W},
  journal={arXiv preprint arXiv:2210.02928},
  year={2022}
}

@inproceedings{van2021benchmarking,
  title={Benchmarking representation learning for natural world image collections},
  author={Van Horn, Grant and Cole, Elijah and Beery, Sara and Wilber, Kimberly and Belongie, Serge and Mac Aodha, Oisin},
  booktitle={Proceedings of the IEEE/CVF conference on computer vision and pattern recognition},
  pages={12884--12893},
  year={2021}
}

@inproceedings{weyand2020google,
  title={Google landmarks dataset v2-a large-scale benchmark for instance-level recognition and retrieval},
  author={Weyand, Tobias and Araujo, Andre and Cao, Bingyi and Sim, Jack},
  booktitle={Proceedings of the IEEE/CVF conference on computer vision and pattern recognition},
  pages={2575--2584},
  year={2020}
}

\appendix
 
\section{Appendix}

\subsection{Details of Baseline Models.}
\label{app-model}
All models evaluated in this work are publicly accessible and obtained from official open sources, including 
\emph{Qwen2.5-VL-Instruct 3B}~\footnote{https://huggingface.co/Qwen/Qwen2.5-VL-3B-Instruct}, \emph{Qwen2.5-VL-Instruct 32B}~\footnote{https://huggingface.co/Qwen/Qwen2.5-VL-32B-Instruct}, \emph{LLaVA-NeXT-Video 34B}~\footnote{https://huggingface.co/llava-hf/LLaVA-NeXT-Video-34B-hf} and \emph{InternVL3 38B}~\footnote{https://huggingface.co/OpenGVLab/InternVL3-38B}.

\subsection{LLM Usage}
We only use the LLMs for correcting the grammar and improving the phrasing during the writing  process.

\subsection{Statistics of VideoSimpleQA}
\begin{table}[h]
\centering
\small
\resizebox{\linewidth}{!}{
\begin{tabular}{lccc}
\toprule
\textbf{\# questions} & {\textbf{\# videos}} & \textbf{Average Video Length(s)} & \textbf{\# documents} \\
\midrule
            203&102&22.27&156 \\
\bottomrule
\end{tabular}}
\caption{Statistics of VideoSimpleQA.}
\end{table}
\label{sec:appendix}

\end{document}